\documentclass[10pt,twocolumn,letterpaper]{article}

\usepackage{cvpr}      

\usepackage[dvipsnames]{xcolor}

\usepackage{times}
\usepackage{epsfig}
\usepackage{graphicx}
\usepackage{amsmath}
\usepackage{amssymb}
\usepackage[numbers,sort&compress]{natbib}

\usepackage[mathscr]{eucal}
\usepackage[ruled,vlined,linesnumbered]{algorithm2e}
\usepackage{graphicx}  
\usepackage{epstopdf}
\usepackage{booktabs}
\usepackage{multirow}
\usepackage{setspace}
\usepackage{epstopdf}

\usepackage{tikz-imagelabels}
\usepackage{pifont}

\def\softmax{{\ttfamily softmax} }
\def\softmaxf{\text{\ttfamily softmax}}
\def\xv{{\mathbf{x}}}
\def\y{{{y}}}

\def\zv{{\mathbf{z}}}
\def\vv{{\mathbf{v}}}
\def\T{{\mathcal{T}}}
\def\D{{\mathcal{D}}}
\def\Loss{{\mathcal{L}}}
\def\zscore{{$\mathcal{Z}$}-score} 
\def\E{{\mathbb{E}}}
\def\KL{{\rm{KL}}}
\def\CE{{\rm{CE}}}

\definecolor{c6}{HTML}{ff0097}

\definecolor{cvprblue}{rgb}{0.21,0.49,0.74}
\usepackage[pagebackref,breaklinks,colorlinks,citecolor=cvprblue]{hyperref}

\def\paperID{1508} 
\def\confName{CVPR}
\def\confYear{2024}

\title{Logit Standardization in Knowledge Distillation}

\author{Shangquan Sun$^{1,2}$, Wenqi Ren$^{3\dag}$, Jingzhi Li$^{1}$, Rui Wang$^{1,2}$, Xiaochun Cao$^{3}$\\
$^{1}$Institute of Information Engineering, Chinese Academy of Sciences \\ 
$^{2}$School of Cyber Security, University of Chinese Academy of Sciences \\ 
$^{3}$School of Cyber Science and Technology, Sun Yat-sen University, Shenzhen Campus \\
{\tt\small \{sunshangquan,lijingzhi,wangrui\}@iie.ac.cn}, {\tt\small \{renwq3,caoxiaochun\}@mail.sysu.edu.cn}
}

\begin{document}
\maketitle

\renewcommand{\thefootnote}{}
\footnotetext{\dag \ Corresponding author.}

\begin{abstract}
   Knowledge distillation involves transferring soft labels from a teacher to a student using a shared temperature-based \softmax function. 
   However, the assumption of a shared temperature between teacher and student implies a mandatory exact match between their logits in terms of logit range and variance.
   This side-effect limits the performance of student, considering the capacity discrepancy between them and the finding that the innate logit relations of teacher are sufficient for student to learn.
   %
   To address this issue, we propose setting the temperature as the weighted standard deviation of logit and performing a plug-and-play {\zscore} pre-process of logit standardization before applying \softmax and Kullback-Leibler divergence. 
   Our pre-process enables student to focus on essential logit relations from teacher rather than requiring a magnitude match, and can improve the performance of existing logit-based distillation methods. 
   We also show a typical case where the conventional setting of sharing temperature between teacher and student cannot reliably yield the authentic distillation evaluation; nonetheless, this challenge is successfully alleviated by our {\zscore}.
   We extensively evaluate our method for various student and teacher models on CIFAR-100 and ImageNet, showing its significant superiority. 
   The vanilla knowledge distillation powered by our pre-process can achieve favorable performance against state-of-the-art methods, and other distillation variants can obtain considerable gain with the assistance of our pre-process.
   The codes, pre-trained models and logs are released on \href{https://github.com/sunshangquan/logit-standardardization-KD}{Github}.
\end{abstract}

\abovedisplayshortskip=1.6pt
\belowdisplayshortskip=1.6pt
\abovedisplayskip=1.6pt
\belowdisplayskip=1.6pt

\vspace{-3mm}
\section{Introduction}

\begin{figure}[t]
\begin{center}
   \subfloat[Vanilla KD]{\includegraphics[width=0.47\linewidth]{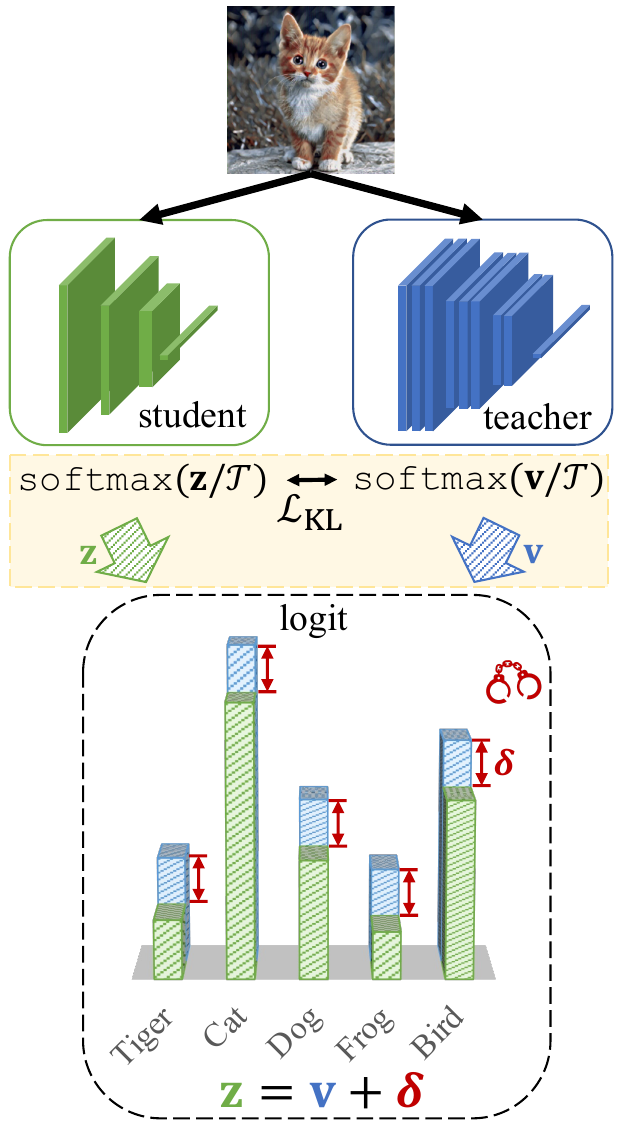}}
   \hfill
   \subfloat[KD w/ our logit standardization]{\includegraphics[width=0.47\linewidth]{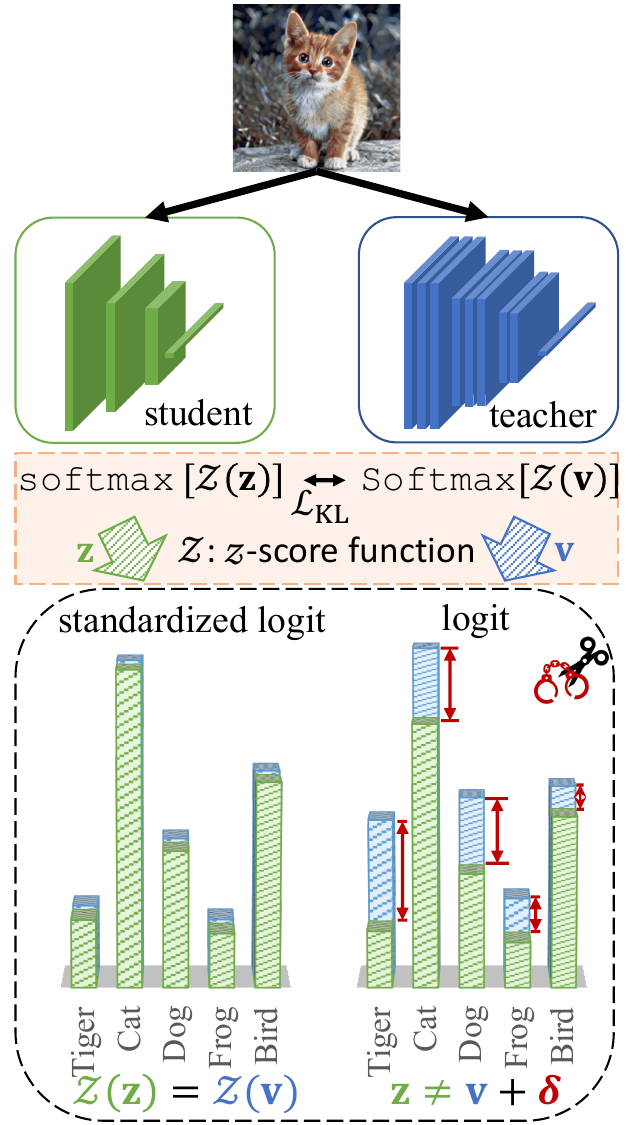}}
\end{center}
   \vspace{-6.5mm}
   \caption{
   Vanilla knowledge distillation implicitly enforces an exact match between the magnitudes of teacher and student logits. 
   It is an unnecessary side-effect because it is found sufficient to preserve the innate relations between their logits. 
   Given the capacity gap between them, it is also challenging for a lightweight student to produce logits with the same magnitude as a cumbersome teacher. 
   In contrast, the proposed {\zscore} logit standardization pre-process mitigates the side-effect. 
   The standardized student logits have arbitrary magnitude suitable for the student's capacity while preserving the essential relations learned from the teacher.
   }
   \vspace{-3.5mm}
\label{fig:cover}
\end{figure}
\vspace{-1.5mm}
The development of deep neural networks (DNN) has revolutionized the field of computer vision in the past decade. However, with increasing performance and capacity, the model size and computational cost of DNN have also been expanding. 
Despite of a tendency that larger models have greater capacity, many efforts have been made by researchers to cut down model size without sacrificing much accuracy. 
In addition to designing lightweight models, knowledge distillation (KD) has emerged as a new approach to achieve this goal. 
It involves transferring the knowledge of a pre-trained heavy model, known as the teacher network, to a small target model, known as the student network. 

Hinton \textit{et al}.~\cite{hinton2015distilling} firstly proposes distilling a teacher's knowledge into a student by minimizing a Kullback-Leibler (KL) divergence between their predictions. 
A scaling factor of the \softmax function in this context, called temperature $\T$, is introduced to soften the predicted probabilities. 
Traditionally, the temperature is set globally beforehand as a hyper-parameter and remains fixed throughout training.
CTKD~\cite{li2022curriculum} adopts an adversarial learning module to predict sample-wise temperatures, adapting to varying sample difficulties. 
However, existing logit-based KD approaches still assume that the teacher and student should share temperatures, neglecting the possibility of distinct temperature values in the KL divergence. 
%
In this work, we demonstrate that the general \softmax expression in both classification and KD is derived from the principle of entropy maximization in information theory. 
During this derivation, Lagrangian multipliers appear and take the form of temperatures, based on which we establish the irrelevance between the temperatures of teacher and student, as well as the irrelevance among temperatures for different samples.
The proof supports our motivation to allocate distinct temperatures between teacher and student and across samples.

%
%
%

Compared to an exact match of logit prediction, it is found that the inter-class relations of predictions are sufficient for student to achieve performance similar to teacher~\cite{huang2022knowledge}. 
A lightweight student faces challenges in predicting logits with a comparable range and variance as a cumbersome teacher, given the capacity gap between them~\cite{cho2019efficacy,mirzadeh2020improved,son2021densely}. 
However, we demonstrate that the conventional practice of sharing temperatures in KL divergence still implicitly enforces an exact match between the student and teacher's logits.
Existing logit-based KD methods, unaware of this issue, commonly fall in the pitfall, resulting in a general performance drop. 
To address this, we propose using weighted logit standard deviation as an adaptive temperature and present a {\zscore} logit standardization as a pre-processing step before applying \softmax. 
This pre-processing maps arbitrary range of logits into a bounded range, allowing student logits to possess arbitrary ranges and variances while efficiently learning and preserving only the innate relationships of teacher logits.
We present a typical case where the KL divergence loss under the setting of sharing temperatures in \softmax may be misleading and cannot reliably measure the performance of distilled students.
In contrast, with our {\zscore} pre-process, the issue of the shared temperatures in the case is eliminated.

In summary, our contributions are in three folds
\begin{itemize}
\setlength{\itemsep}{0pt}
\setlength{\parsep}{0pt}
\setlength{\parskip}{0pt}
    \item Based on the entropy maximization principle in information theory, we use Lagrangian multipliers to derive the general expression of \softmax in logit-based KD. 
    We show that the temperature comes from the derived multipliers, allowing it to be selected differently for various samples and distinctly for student and teacher.
    \item To address the issues of the conventional logit-based KD pipeline caused by shared temperatures, including an implicit mandatory logit match and an inauthentic indication of student performance, we propose a pre-process of logit distillation to adaptively allocate temperatures between teacher and student and across samples, capable of facilitating existing logit-based KD approaches. 
    \item We conduct extensive experiments with various teacher and student models on CIFAR-100~\cite{krizhevsky2009learning} and ImageNet~\cite{russakovsky2015imagenet} and demonstrate the superior advantages of our method as a plug-and-play pre-process.
\end{itemize}

\vspace{-1.5mm}
\section{Related Work}
\vspace{-1.5mm}

Knowledge distillation~\cite{hinton2015distilling} is designed to transfer the ``dark'' knowledge from a cumbersome teacher model to a lightweight student model. 
By learning from the soft labels of teacher, student can achieve better performance than training on hard labels only. 
The traditional method trains a student by minimizing a difference such as KL divergence between its predicted probability and the teacher's. 
The prediction of probability is commonly approximated by the \softmax of logit output. 
KD algorithms can be classified into three types, i.e.,
logit-based~\cite{chen2020online,hinton2015distilling,li2020online,mirzadeh2020improved,zhang2018deep,zhao2022decoupled,jin2023multi,zhang2023blindly}, 
feature-based~\cite{chen2022knowledge,chen2021distilling,heo2019comprehensive,lin2022knowledge,romero2014fitnets,tian2019contrastive,yim2017gift,ahn2019variational,li2021online,liu2019structured,guo2023class},
and relation-based~\cite{park2019relational,peng2019correlation,li2022knowledge,tung2019similarity,yang2021knowledge,huang2022knowledge} methods.

A temperature is introduced to flatten the probabilities in logit-based methods. 
Several works~\cite{chandrasegaran2022revisiting,hinton2015distilling,liu2022meta} explore its properties and effects. 
They reach an identical conclusion that temperature controls how much attention student pays on those logits more negative than average. 
A very low temperature makes student ignore other logits and instead mainly focus on the largest logit of teacher. 
However, they do not discuss why teacher and student share a globally predefined temperature. 
It was unknown whether temperature can be determined in an instance-wise level until CTKD~\cite{li2022curriculum} proposed predicting sample-wise temperatures by leveraging adversarial learning. 
However, it assumes that teacher and student should share temperatures.
It was still undiscovered whether teacher and student can have divergent temperatures. 
ATKD~\cite{guo2022reducing} proposes a sharpness metric and chooses adaptive temperature by reducing the gap between teacher and student. 
However, their assumption of a zero logit mean relies on numerical approximation and limits its performance. 
Additionally, they do not thoroughly discuss where the temperature is derived from and whether distinct temperatures can be assigned.
In this work, we provide an analytical derivation based on the entropy-maximization principle, demonstrating that students and teachers do not necessarily share a temperature. 
It is also found sufficient to preserve the innate relationship of prediction, instead of exact logit values of teacher~\cite{huang2022knowledge}.
However, the existing logit-based KD pipelines still implicitly mandate an exact match between teacher and student logits.
We thus define the temperature to be the weighted standard deviation of logit
to alleviate the issue and facilitate the existing logit-based KD approaches.





\vspace{-1.5mm}
\section{Background and Notation}
\vspace{-1.5mm}
Suppose we have a transfer dataset $\D$ containing totally $N$ samples $\{\xv_n,\y_n\}_{n=1}^N$, 
where $\xv_n\in \mathbb{R}^{H\times W}$ and $\y_n\in [1,K]$ are the image and label respectively for the $n$-th sample. 
The notations of $H$, $W$ and $K$ are image height, width and the number of classes. 
Given an input $\{\xv_n,\y_n\}$, teacher $f_T$ and student $f_S$ respectively predict logit vectors $\vv_n$ and $\zv_n\in \mathbb{R}^{1\times K}$. 
Namely, $\zv_n=f_S(\xv_n)$ and $\vv_n=f_T(\xv_n)$.

It is widely accepted that a \softmax function involving a temperature $\T$ is used to convert the logit to probability vectors $q(\zv_n)$ or $q(\vv_n)$ such that their $k$-th items have
\begin{align}
    q(\zv_n)^{(k)} &= \frac{\exp(\zv_n^{(k)}/\T)}{\sum_{m=1}^K \exp(\zv_n^{(m)}/\T)}\label{eq::sharedProbStu}, \\
    q(\vv_n)^{(k)} &= \frac{\exp(\vv_n^{(k)}/\T)}{\sum_{m=1}^K \exp(\vv_n^{(m)}/\T)}\label{eq::sharedProbTea},
\end{align}
where $\zv_n^{(k)}$ and $\vv_n^{(k)}$ are the $k$-th item of $\zv_n$ and $\vv_n$ respectively. 
A knowledge distillation process is essentially letting $q(\zv_n)^{(k)}$ mimic $q(\vv_n)^{(k)}$ for any class and all samples. 
The objective is realized by minimizing KL divergence
\begin{equation*}\label{eq::KL}
    \Loss_\KL\left(q(\vv_n)||q(\zv_n)\right) = \sum_{k=1}^K q(\vv_n)^{(k)} \log \left(\frac{q(\vv_n)^{(k)}}{q(\zv_n)^{(k)}}\right),
\end{equation*}
\noindent which is theoretically equivalent to a cross-entropy loss when optimizing solely on $\zv$,
\begin{equation}\label{eq::CE}
    \Loss_\CE\left(q(\vv_n),q(\zv_n)\right) = -\sum_{k=1}^K q(\vv_n)^{(k)} \log {q(\zv_n)^{(k)}}.
\end{equation}
Note that they are empirically nonequivalent as their gradients diverge due to the negative entropy term of $q(\vv_n)$.

\vspace{-1.5mm}
\section{Methodology}
\vspace{-1.5mm}

It is widely accepted that $\T$ is shared for teacher and student in Eq.~\ref{eq::sharedProbStu} and Eq.~\ref{eq::sharedProbTea}. 
In contrast, in Sec.~\ref{sec::irrelevance}, we show the irrelevance between the temperatures of teacher and student, as well as across different samples.
Guaranteed that temperatures can be different between teacher and student and among sample, we further show two side-effect drawbacks of shared-temperatures setting in conventional KD pipelines in Sec.~\ref{sec::drawbacks}.
In Sec.~\ref{sec::logit_standard}, we propose leveraging logit standard deviation as a factor in temperature and derive a pre-process of logit standardization. 

\vspace{-1.5mm}
\subsection{Irrelevance between Temperatures}
\label{sec::irrelevance}
\vspace{-1.5mm}
In Sec.~\ref{sec::softmaxCl} and \ref{sec::softmaxKD}, we first give a derivation of the temperature-involved \softmax function in classification and KD based on the entropy-maximization principle in information theory. 
This implies the temperatures of student and teacher can be distinct and sample-wisely different. 

\vspace{-4.5mm}
\subsubsection{Derivation of \softmax in Classification}\label{sec::softmaxCl}
\vspace{-1.5mm}

The \softmax function in classification can be proved to be the unique solution of maximizing entropy subject to the normalization condition of probability and a constraint on the expectation of states in information theory~\cite{jaynes1957information}. 
The derivation is also leveraged in confidence calibration to formulate temperature scaling~\cite{guo2017calibration}. 
Suppose we have the following constrained entropy-maximization optimization, 
%
%
\begin{equation}\label{eq::objCL}
{\small
\begin{split}
    &\max_{q} \mathcal{L}_1= -\sum_{n=1}^N\sum_{k=1}^K q(\vv_n)^{(k)} \log q(\vv_n)^{(k)} \\
    \textit{s.t. } &\left\{
    \begin{aligned}
        &\sum_{k=1}^K q(\vv_n)^{(k)} = 1, \quad \forall n \\
        & \E_q[\vv_n] = \sum_{k=1}^K \vv_n^{(k)}q(\vv_n)^{(k)} = \vv_n^{(\y_n)}, \quad \forall n.
    \end{aligned} \right.
\end{split}
}
\end{equation}
The first constraint holds due to the requirement of discrete probability density, while the second constraint controls the scope of the distribution such that model accurately predicts the target class. 
Suppose $\hat q_n$ to be the one-hot hard probability distribution whose values are all zero except at the target index $\hat q_n^{(\y_n)}=1$.
The second constraint is then actually $\E_q[\vv_n] = \sum_{k=1}^K \vv_n^{(k)}\hat q_n^{(k)} = \vv_n^{(y_n)}$. 
This is equivalent to making model predict the correct label $\y_n$.
By applying Lagrangian multipliers $\{\alpha_{1,i}\}_{i=1}^N$ and $\{\alpha_{2,i}\}_{i=1}^N$, it gives
\begin{equation*}
\begin{split}
    \Loss_T = \mathcal{L}_1
    & + \sum_{n=1}^N \alpha_{1,n} \left(\sum_{k=1}^K q(\vv_n)^{(k)} - 1 \right) \\
    & + \sum_{n=1}^N \alpha_{2,n} \left(\sum_{k=1}^K \vv_n^{(k)}q(\vv_n)^{(k)} - \vv_n^{(\y_n)}\right).
\end{split}
\end{equation*}
Taking the partial derivative with respective to $\alpha_{1,n}$ and $\alpha_{2,n}$ yields back the constraints. 
In contrast, taking the derivative with respective to $q(\vv_n)^{(k)}$ gives
\begin{equation}
    \frac{\partial \Loss_T}{\partial q(\vv_n)^{(k)}} = -1 - \log q(\vv_n)^{(k)} + \alpha_{1,n} + \alpha_{2,n}\vv_n^{(k)},
\end{equation}
which leads to a solution by making the derivative zero: 
\begin{equation}\label{eq::teacher_prob}
    q(\vv_n)^{(k)} = 
    \exp \left(\alpha_{2,n}\vv_n^{(k)}\right) / Z_T,
\end{equation}
where $Z_T=\exp \left(1 - \alpha_{1,n} \right) = \sum_{m=1}^K \exp\left(\alpha_{2,n}\vv_n^{(m)}\right)$ is the partition function to fulfill the normalization condition. 

\vspace{-4.5mm}
\subsubsection{Derivation of \softmax in KD}
\label{sec::softmaxKD}
\vspace{-1.5mm}

Following the idea, we define a problem of entropy-maximization to formulate the \softmax in KD. 
Given a well-trained teacher and its prediction $q(\vv_n)$, we have the objective function for the prediction of student as follows,
\begin{equation}\label{eq::objKD}
{\small
\begin{split}
    &\max_{q} \mathcal{L}_2= -\sum_{n=1}^N\sum_{k=1}^K q(\zv_n)^{(k)} \log q(\zv_n)^{(k)} \\
    \textit{s.t. } &\left\{
    \begin{split}
        & \sum_{k=1}^K q(\zv_n)^{(k)} = 1,\quad \forall n \\
        & \sum_{k=1}^K \zv_n^{(k)}q(\zv_n)^{(k)} = \zv_n^{(\y_n)},\quad \forall n \\
        & \sum_{k=1}^K \zv_n^{(k)}q(\zv_n)^{(k)} = \sum_{k=1}^K \zv_n^{(k)}q(\vv_n)^{(k)},\quad \forall n.
    \end{split} \right.
\end{split}
}
\end{equation}
By applying Lagrangian multipliers $\beta_{1,n}$, $\beta_{2,n}$ and $\beta_{3,n}$, 
\begin{equation*}
\begin{split}
    \Loss_S = \mathcal{L}_2
    & + \sum_{n=1}^N \beta_{1,n}\left(\sum_{k=1}^K q(\zv_n)^{(k)} - 1\right) \\
    & + \sum_{n=1}^N \beta_{2,n} \left(\sum_{k=1}^K \zv_n^{(k)}q(\zv_n)^{(k)} - \zv_n^{(\y_n)} \right) \\
    & + \sum_{n=1}^N \beta_{3,n} \sum_{k=1}^K \zv_n^{(k)} \left(q(\zv_n)^{(k)} - q(\vv_n)^{(k)}\right).
\end{split}
\end{equation*}
Taking its derivative with respective to $q(\zv_n)^{(k)}$ gives
\begin{equation*}
    \frac{\partial \Loss_S}{\partial q(\zv_n)^{(k)}} = -1 - \log q(\zv_n)^{(k)} + \beta_{1,n} + \beta_{2,n}\zv_n^{(k)} + \beta_{3,n} \zv_n^{(k)}.
\end{equation*}
Suppose $\beta_n=\beta_{2,n}+\beta_{3,n}$ for simplicity and it gives 
\begin{equation}\label{eq::student_prob}
    q(\zv_n)^{(k)} = 
    \exp \left(\beta_n\zv_n^{(k)}\right) / Z_S,
\end{equation}
where $Z_S = \exp(1-\beta_{1,n}) = \sum_{k=1}^K \exp(\beta_n \zv_n^{(k)})$ holds because of the normalization condition of probability density. The formulation in Eq.~\ref{eq::student_prob} has the same structure as Eq.~\ref{eq::teacher_prob}. 

\noindent\textbf{Distinct Temperatures.} 
Note that the partial derivatives of $\Loss_T $ with respective to $\alpha_{1,n}$ and $\alpha_{2,n}$ lead back to the two constraints respectively in Eq.~\ref{eq::objCL} and the constraints are irrelevant to $\alpha_{1,n}$ and $\alpha_{2,n}$. 
A similar case holds for Eq.~\ref{eq::objKD}. 
As a result, no explicit expression of them can be given and thus their values can be manually defined. 
If we set $\beta_n=\alpha_{2,n}=1/\T$, Eq.~\ref{eq::teacher_prob} and \ref{eq::student_prob} turn to the expressions in KD involving a shared temperature for both student and teacher. 
When $\beta_n=\alpha_{2,n}=1$, the formulations revert to the traditional \softmax function commonly used in classification. 
Eventually, we can choose $\beta_n\ne \alpha_{2,n}$, indicating that students and teachers can have distinct temperatures.

\noindent\textbf{Sample-wisely different Temperatures.} 
It is common to define a global temperature for all samples. 
Namely for any $n$, $\alpha_{2,n}$ and $\beta_n$ are defined as a constant value. 
In contrast, they could vary across different samples due to the lack of restrictions on them. 
It lacks a foundation to choose a global constant value as temperature. 
As a result, adopting sample-wise varying temperatures is allowed.

\vspace{-1.5mm}
\subsection{Drawbacks of Shared Temperatures}
\label{sec::drawbacks}
\vspace{-1.5mm}
In this section, we show two shortcomings of the shared temperatures setting in conventional KD pipeline.
We first rewrite the \softmax in Eq.~\ref{eq::student_prob} in a general formulation by introducing two hyper-parameters $a_S$ and $b_S$,
\begin{equation*}
    q\left(\zv_n; a_S, b_S\right)^{(k)} = \frac{\exp\left[(\zv_n^{(k)}-a_S)/b_S\right]}{\sum_{m=1}^K \exp\left[(\zv_n^{(m)}-a_S)/b_S\right]},
\end{equation*}
where $a_S$ can be cancelled out and does not violate the equality. 
When $a_S=0,b_S=1/\beta_n$, it yields back the special case in Eq.~\ref{eq::student_prob}. 
The similar equation for the case of teacher can be written by introducing $a_T$ and $b_T$. 

For a finally well-distilled student, we assume the KL divergence loss reaches minimum and its predicted probability density matches that of teacher, i.e., $\forall k\in [1,K]$, $q(\zv_n; a_S, b_S)^{(k)} = q(\vv_n; a_T, b_T)^{(k)}$. 
Then for arbitrary pair of indices $i,j\in [1,K]$, it can easily lead to 
\begin{align*}
    &\frac{\exp\left[(\zv_n^{(i)}-a_S)/b_S\right]}{\exp\left[(\zv_n^{(j)}-a_S)/b_S\right]} = \frac{\exp\left[(\vv_n^{(i)}-a_T)/b_T\right]}{\exp\left[(\vv_n^{(j)}-a_T)/b_T\right]} \\
    \Rightarrow & \left(\zv_n^{(i)} - \zv_n^{(j)}\right) / b_S = \left(\vv_n^{(i)} - \vv_n^{(j)}\right) / b_T.
\end{align*}
By taking a summation across $j$ from $1$ to $K$, we have
\begin{equation}\label{eq::logitShift}
    \left( \zv_n^{(i)} - \overline{\zv}_n \right) / b_S = \left( \vv_n^{(i)} - \overline{\vv}_n \right) / b_T,
\end{equation}
where $\overline{\zv}_n$ and $\overline{\vv}_n$ are the mean of the student and teacher logit vectors respectively, i.e., $\overline{\zv}_n=\frac{1}{K}\sum_{m=1}^K\zv_n^{(m)}$ (similar and omitted for $\overline{\vv}_n$). 
By taking the summation of the squared Eq.~\ref{eq::logitShift} across $i$ from $1$ to $K$, we can obtain
\begin{equation}\label{eq::logitRatio}
    \frac{\sigma(\zv_n)^2}{\sigma(\vv_n)^2} = \frac{\frac{1}{K}\sum_{i=1}^K\left( \zv_n^{(i)} - \overline{\zv}_n \right)^2 }{\frac{1}{K}\sum_{i=1}^K \left( \vv_n^{(i)} - \overline{\vv}_n \right)^2} = \frac{b_S^2 }{ b_T^2},
\end{equation}
where $\sigma$ is the function of standard deviation for an input vector. 
From Eq.\ref{eq::logitShift} and \ref{eq::logitRatio}, we can describe two properties of a well-distilled student in terms of the logit shift and variance matching.

\noindent\textbf{Logit shift.} 
From Eq.\ref{eq::logitShift}, it can be found that a constant shift exists between the logits of student and teacher in arbitrary index under the traditional setting of shared temperature ($b_S=b_T$), i.e.,
\begin{equation}\label{eq::shackle1}
    \zv_n^{(i)} = \vv_n^{(i)} + \Delta_n,
\end{equation}
where $\Delta_n = \overline{\zv}_n - \overline{\vv}_n$ can be considered as constant for the $n$-th sample. 
This implies in the traditional KD approach, student is forced to strictly mimic the shifted logit of teachers. 
Considering the gap of their model size and capacity, student may be unable to produce as wide logit range as teacher~\cite{cho2019efficacy,mirzadeh2020improved,son2021densely}. 
In contrast, a student can be considered excellent enough when its logit rank matches teacher~\cite{huang2022knowledge}, i.e., given the indices that sorts the teacher logits $t_1,...,t_K\in[1,K]$ such that $\vv_n^{(t_1)} \le \cdots \le\vv_n^{(t_K)}$, then $\zv_n^{(t_1)} \le \cdots \le\zv_n^{(t_K)}$ holds. 
The logit relation is the essential knowledge that makes student predict as excellently as teacher. 
Such a logit shift is thus a side-effect in conventional KD pipeline and a shackle compelling student to generate unnecessarily difficult results.

\noindent\textbf{Variance match.} 
From Eq.~\ref{eq::logitRatio}, we come into a conclusion that the ratio between the temperatures of student and teacher equals the ratio between the standard deviations of their predicted logits for a well-distilled student, i.e.,
\begin{equation}\label{eq::shackle2}
    \sigma(\zv_n)/\sigma(\vv_n) = b_S/b_T.
\end{equation}
In the setting of temperature sharing in vanilla KD, the student is forced to predict logit such that $\sigma(\zv_n)=\sigma(\vv_n)$. 
This is another shackle applied to student restricting the standard deviation of its predicted logits. 
In contrast, since the hyper-parameter comes from Lagrangian multiplier and is flexible to tune, we can define $b_S^* \propto \sigma(\zv_n)$ and $b_T^*\propto \sigma(\vv_n)$. 
In that way, the equation in Eq.~\ref{eq::shackle2} always holds.

\setlength{\textfloatsep}{0.3cm}
\setlength{\floatsep}{0.0cm}
\begin{algorithm}[t]
    \SetAlgoLined
	\caption{Weighted {\zscore} function.}\label{alg::zscoref}
	\KwIn{Input vector $\xv$ and Base temperature $\tau$}
	\KwOut{Standardized vector $\mathcal{Z} (\xv; \tau)$}  
	\BlankLine
	
	

            $\overline{\xv} \leftarrow \frac{1}{K} \sum_{k=1}^K \xv^{(k)}$
            
            $\sigma (\xv) \leftarrow \sqrt{{\frac{1}{K} \sum_{k=1}^K \left(\xv^{(k)} - \overline{\xv}\right)^2}}$
            
        \Return  $(\xv-\overline{\xv} ) / \sigma (\xv) /\tau$
\end{algorithm}

\begin{algorithm}[t]
    \SetAlgoLined
	\caption{{\zscore} logit standardization pre-process in knowledge distillation.}\label{alg::zscore}
	\KwIn{Transfer set $\D$ with image-label sample pair $\{\xv_n,\y_n\}_{n=1}^N$, Base Temperature $\tau$, Teacher $f_T$, Student $f_S$, Loss $\Loss_{\rm KD}$ (e.g., $\Loss_\KL$), loss weight $\lambda$, and {\zscore} function $\mathcal{Z}$ in Algo.~\ref{alg::zscoref}}
	\KwOut{Trained student model $f_S$}  
	\BlankLine
	
	
    
	\ForEach{ $(\xv_n,\y_n)$ in $\D$}{
		$\vv_n \leftarrow f_T(\xv_n)$,\ 
            $\zv_n \leftarrow f_S(\xv_n)$

            $q(\vv_n) \leftarrow \text{\softmaxf} \left[ \mathcal{Z}(\vv_n; \tau) \right]$

            $q(\zv_n) \leftarrow \text{\softmaxf} \left[ \mathcal{Z}(\zv_n; \tau) \right]$

            $q'(\zv_n) \leftarrow \text{\softmaxf} \left( \zv_n\right)$
            

            Update $f_S$ towards minimizing {\footnotesize $\lambda_{\rm CE}\Loss_{\rm CE}\left(\y_n, q'(\zv_n)\right) + \lambda_{\rm KD}\tau^2 \Loss \left(q(\vv_n), q(\zv_n)\right)$}
            
	}
\end{algorithm}

\vspace{-1.5mm}
\subsection{Logit Standardization}\label{sec::logit_standard}
\vspace{-1.5mm}

To break the two shackles, we therefore propose setting the hyper-parameters $a_S$, $b_S$, $a_T$ and $b_S$ to be the mean and the weighted standard deviation of their logits respectively, i.e., 
\begin{equation*}
    q\left(\zv_n; \overline{\zv}_n, \sigma(\zv_n)\right)^{(k)} = \frac{\exp(\mathcal{Z} (\zv_n; \tau)^{(k)})}{\sum_{m=1}^K \exp(\mathcal{Z} (\zv_n; \tau)^{(m)})},
\end{equation*}
%
where $\mathcal{Z}$ is the {\zscore} function in Algo.~\ref{alg::zscoref}. 
The case for teacher logit is similar and omitted. 
A base temperature $\tau$ is introduced and shared for both teacher and student models. 
The {\zscore} standardization has at least four advantageous properties, i.e., zero mean, finite standard deviation, monotonicity and boundedness.

\noindent\textbf{Zero mean.} 
The mean of standardized vector can be easily shown to be zero. 
In previous works~\cite{hinton2015distilling,guo2022reducing}, a zero mean is assumed and usually empirically violated. 
In contrast, the {\zscore} function intrinsically guarantees a zero mean.

\noindent\textbf{Finite standard deviation.} 
The standard deviation of the weighted {\zscore} output $\mathcal{Z} (\zv_n; \tau)$ can be shown equal to $1/\tau$. 
The property makes the standardized student and teacher logit map to an identical Gaussian-like distribution with zero mean and definite standard deviation. 
The mapping of standardization is many-to-one, meaning that its reverse is indefinite. 
The variance and value range of original student logit vector $\zv_n$ is thereby free of restriction.

\noindent\textbf{Monotonicity.} 
It is easy to show that {\zscore} is a linear transformation function and thus lies in monotonic functions. 
This property ensures that the transformed student logit remains the same rank as the original one. 
The necessary innate relation within teacher logit can thus be preserved and transferred to student.

\noindent\textbf{Boundedness.} 
The standardized logit can be shown bounded within $[-\sqrt{K-1}/\tau, \sqrt{K-1}/\tau]$. 
%
Compared to traditional KD, it is feasible to control the logit range and avoid extremely large exponential value. 
To this end, we define a base temperature to control the range. 

The pseudo-code of the proposed logit standardization pre-process is illustrated in Algo.~\ref{alg::zscore}.

\begin{figure}[t]
\begin{center}
   \subfloat{\includegraphics[width=1\linewidth]{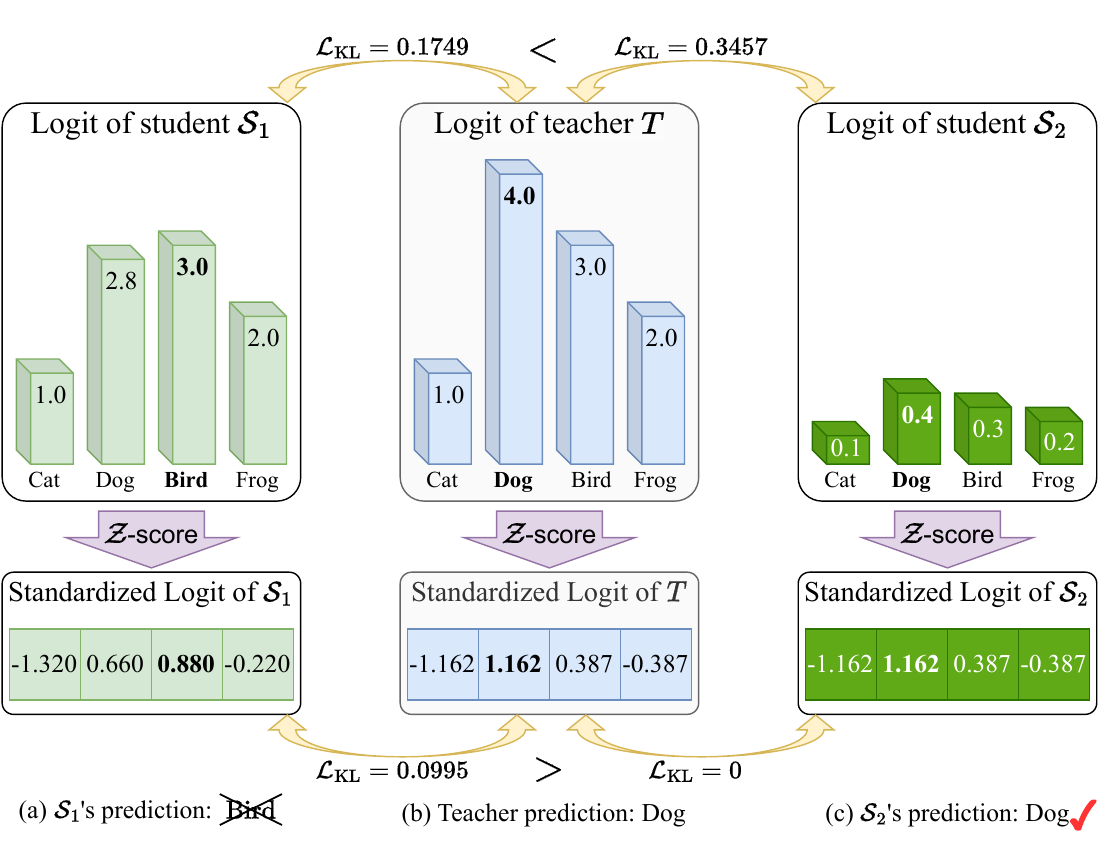}}
\end{center}
   \vspace{-8mm}
   \caption{A toy case where two students, \(\mathcal{S}_1\) and \(\mathcal{S}_2\), learning from the same teacher with an identical temperature (assumed 1 for simplicity). 
   Student \(\mathcal{S}_1\) generates the logits much closer to the teacher's in terms of magnitude and thus has lower loss of 0.1749, but it returns a wrong prediction of ``bird''. 
   In contrast, Student \(\mathcal{S}_2\) outputs the logits far from the teacher's and yields greater loss value of 0.3457, but it returns the correct prediction of ``dog''. 
   After the proposed logit standardization, the issue is addressed.}
\label{fig:toy_case}
\end{figure}

\begin{table*}[htbp]
\footnotesize
\tabcolsep=0.8mm
\renewcommand\arraystretch{0.9} 
  \centering
  \caption{The Top-1 Accuracy (\%) of different knowledge distillation methods on the validation set of CIFAR-100~\cite{krizhevsky2009learning}. The teacher and student have distinct architectures. The KD methods are sorted by the types, i.e., feature-based and logit-based. We apply our logit standardization to the existing logit-based methods and use $\Delta$ to show its performance gain. The values in {\color{blue}blue} denote slight enhancement and those in {\color{red}red} non-trivial enhancement no less than \(0.15\). The best and second best results are emphasized in \textbf{bold} and \(\underline{\text{underlined}}\) cases.}
  \vspace{-2mm}
    \begin{tabular}{llccccccc}
    \toprule
    \multirow{4}{*}{Type} & \multirow{2}{*}{Teacher} & \makebox[0.1\textwidth][c]{ResNet32$\times$4} & \makebox[0.1\textwidth][c]{ResNet32$\times$4} & \makebox[0.1\textwidth][c]{ResNet32$\times$4} & 
    \makebox[0.1\textwidth][c]{WRN-40-2} & \makebox[0.1\textwidth][c]{WRN-40-2} & \makebox[0.1\textwidth][c]{VGG13} &   \makebox[0.1\textwidth][c]{ResNet50} \\
          &       & 79.42 & 79.42 & 79.42 & 75.61 & 75.61 & 74.64 & 79.34 \\
          & \multirow{2}{*}{Student} & SHN-V2 & WRN-16-2 & WRN-40-2 & ResNet8$\times$4 & MN-V2 & MN-V2 & MN-V2 \\
          &       & 71.82 & 73.26 & 75.61 & 72.50 & 64.60 & 64.60 & 64.60 \\
    \midrule
    \multirow{8}{*}{Feature} & FitNet~\cite{romero2014fitnets} & 73.54 & 74.70 & 77.69 & 74.61 & 68.64 & 64.16 & 63.16 \\
          & AT~\cite{zagoruyko2017paying} & 72.73 & 73.91 & 77.43 & 74.11 & 60.78 & 59.40 & 58.58 \\
          & RKD~\cite{park2019relational}   & 73.21 & 74.86 & 77.82 & 75.26 & 69.27 & 64.52 & 64.43 \\
          & CRD~\cite{tian2019contrastive}   & 75.65 & 75.65 & 78.15 & 75.24 & 70.28 & 69.73 & 69.11 \\
          & OFD~\cite{heo2019comprehensive}   & 76.82 & 76.17 & 79.25 & 74.36 & 69.92 & 69.48 & 69.04 \\
          & ReviewKD~\cite{chen2021distilling} & 77.78 & 76.11 & 78.96 & 74.34 & \underline{71.28} & 70.37 & 69.89 \\
          & SimKD~\cite{chen2022knowledge} & 78.39 & \underline{77.17} & \underline{79.29} & 75.29 & 70.10 & 69.44 & 69.97 \\
          & CAT-KD~\cite{guo2023class} & 78.41 & 76.97 & 78.59 & 75.38 & 70.24 & 69.13 &  \textbf{71.36} \\
    \midrule
    \multirow{13}{*}{Logit} & KD~\cite{hinton2015distilling}    & 74.45 & 74.90 & 77.70 & 73.97 & 68.36 & 67.37 & 67.35 \\
          & KD+Ours & 75.56 & 75.26 & 77.92 & 77.11 & 69.23 & 68.61 & 69.02 \\
          & $\Delta$ & {\color{red}1.11}  & {\color{red}0.36}  & {\color{red}0.22}  & {\color{red}3.14}  & {\color{red}0.87}  & {\color{red}1.24}  & {\color{red}1.67} \\
\cmidrule{2-9}          & CTKD~\cite{li2022curriculum}  & 75.37 & 74.57 & 77.66 & 74.61 & 68.34 & 68.50 & 68.67 \\
          & CTKD+Ours & 76.18 & 75.16 & 77.99 & 77.03 & 69.53 & 68.98 & 69.36  \\
          & $\Delta$ & {\color{red}0.81}  & {\color{red}0.59}  & {\color{red}0.33}  & {\color{red}2.42}  & {\color{red}1.19}  & {\color{red}0.48}  & {\color{red}0.69} \\
\cmidrule{2-9}          & DKD~\cite{zhao2022decoupled}   & 77.07 & 75.70 & 78.46 & 75.56 & 69.28 & 69.71 & 70.35 \\
          & DKD+Ours & 77.37 & 76.19 & 78.95 & 76.75 & 70.01 & 69.98 & 70.45 \\
          & $\Delta$ & {\color{red}0.30}  & {\color{red}0.49}  & {\color{red}0.49}  & {\color{red}1.19}  & {\color{red}0.73}  & {\color{red}0.27}  & {\color{blue}0.10} \\
\cmidrule{2-9}          & MLKD~\cite{jin2023multi}   & \underline{78.44} & 76.52 & 79.26  & \underline{77.33} & 70.78 &  \underline{70.57} & 71.04 \\
          & MLKD+Ours & \textbf{78.76} & \textbf{77.53} & \textbf{79.66}  & \textbf{77.68} & \textbf{71.61} & \textbf{70.94} & \underline{71.19} \\
          & $\Delta$ & {\color{red}0.32} & {\color{red}1.01} & {\color{red} 0.40} & {\color{red}0.35} & {\color{red}0.83} & {\color{red}0.37} & {\color{red}0.15} \\ 
    \bottomrule
    \end{tabular}%
  \label{tab:diffStru}%
\vspace{-4mm}
\end{table*}%

\vspace{-4mm}
\subsubsection{Toy Case}
\label{sec::toy_case}
\vspace{-1.5mm}
Fig.~\ref{fig:toy_case} shows a typical case where the conventional logit-based KD setting of shared temperature may lead to an inauthentic evaluation of student performance. 
The first student $\mathcal{S}_1$ predicts logits closer to the teacher $T$ in terms of magnitude, while the second student $\mathcal{S}_2$ preserves the same innate logit relations as the teacher.
As a result, $\mathcal{S}_1$ obtains lower KL divergence loss of 0.1749, much better than the second student $\mathcal{S}_2$ (0.3457). 
However, $\mathcal{S}_1$ has a wrong prediction of ``Bird'' while $\mathcal{S}_2$ predicts ``Dog'' correctly, which contradicts the loss comparison. 
By applying our {\zscore}, all logits are re-scaled and the relations among logits instead of their magnitudes are emphasized in the evaluation. 
Namely, $\mathcal{S}_2$ gets a loss of $0$, much better than $\mathcal{S}_1$ of 0.0995, which is in line with the observed predictions.

\vspace{-2.5mm}
\section{Experiments}
\vspace{-1.5mm}
\noindent\textbf{Datasets.} 
We conduct experiments on CIFAR-100~\cite{krizhevsky2009learning} and ImageNet~\cite{russakovsky2015imagenet}. CIFAR-100~\cite{krizhevsky2009learning} is a common dataset for image classification consisting of 50,000 training and 10,000 validation images. It contains 100 classes and its image size is $32\times 32$. ImageNet~\cite{russakovsky2015imagenet} is a large-scale dataset for image classification containing 1,000 categories and around 1.28 million training and 50,000 validation images.

\noindent\textbf{Baselines.} We evaluate the effect of our logit standardization as pre-process for multiple logit-based KD approaches, including KD~\cite{hinton2015distilling}, CTKD~\cite{li2022curriculum}, DKD~\cite{zhao2022decoupled} and MLKD\cite{jin2023multi}. 
We also compare with various feature-based KD methods including FitNet~\cite{romero2014fitnets}, RKD~\cite{park2019relational}, CRD~\cite{peng2019correlation}, OFD~\cite{heo2019comprehensive}, ReviewKD~\cite{chen2021distilling}, SimKD~\cite{chen2022knowledge}, and CAT-KD~\cite{guo2023class}.

\begin{table*}[htbp]
\footnotesize
\tabcolsep=0.8mm
\renewcommand\arraystretch{0.9} 
  \centering
  \caption{The Top-1 Accuracy (\%) of different knowledge distillation methods on the validation set of CIFAR-100~\cite{krizhevsky2009learning}. The teacher and student have identical architectures but different configurations. The KD methods are sorted by the types. We apply our logit standardization to the existing logit-based methods and use $\Delta$ to show its performance gain. The values in {\color{blue}blue} denote slight enhancement and those in {\color{red}red} non-trivial enhancement no less than \(0.15\). The best and second best results are emphasized in \textbf{bold} and \underline{underlined} cases.}
  \vspace{-2mm}
    \begin{tabular}{llccccccc}
    \toprule
    \multirow{4}{*}{Type} & \multirow{2}{*}{Teacher} & \makebox[0.1\textwidth][c]{ResNet32$\times$4} & \makebox[0.1\textwidth][c]{VGG13} &   \makebox[0.1\textwidth][c]{WRN-40-2} & \makebox[0.1\textwidth][c]{WRN-40-2} & \makebox[0.1\textwidth][c]{ResNet56} & \makebox[0.1\textwidth][c]{ResNet110} & \makebox[0.1\textwidth][c]{ResNet110} \\
          &  & 79.42     & 74.64 & 75.61 & 75.61 & 72.34 & 74.31 & 74.31  \\
          & \multirow{2}{*}{Student} & ResNet8$\times$4 & VGG8  & WRN-40-1 & WRN-16-2 & ResNet20 & ResNet32 & ResNet20 \\
          & & 72.50      & 70.36 & 71.98 & 73.26 & 69.06 & 71.14 & 69.06  \\
    \midrule
    \multirow{8}{*}{Feature} & FitNet~\cite{romero2014fitnets} & 73.50 & 71.02 & 72.24 & 73.58 & 69.21 & 71.06 & 68.99 \\
          & AT~\cite{zagoruyko2017paying} & 73.44 & 71.43 & 72.77 & 74.08 &  70.55 & 72.31 & 70.65  \\
          & RKD~\cite{park2019relational} & 71.90   & 71.48 & 72.22 & 73.35 & 69.61 & 71.82 & 69.25 \\
          & CRD~\cite{tian2019contrastive} & 75.51   & 73.94 & 74.14 & 75.48 & 71.16 & 73.48 & 71.46 \\
          & OFD~\cite{heo2019comprehensive} & 74.95   & 73.95 & 74.33 & 75.24 & 70.98 & 73.23 &   71.29    \\
          & ReviewKD~\cite{chen2021distilling} & 75.63 & 74.84 & 75.09 & 76.12 & 71.89 & 73.89 &    71.34   \\
          & SimKD~\cite{chen2022knowledge} & \underline{78.08} &   74.89    & 74.53 &  75.53     & 71.05      &  73.92  &  71.06   \\
          & CAT-KD~\cite{guo2023class} & 76.91 & 74.65 & 74.82 & 75.60 & 71.62 & 73.62 & 71.37  \\
    \midrule
    \multirow{13}{*}{Logit} & KD~\cite{hinton2015distilling} & 73.33    & 72.98 & 73.54 & 74.92 & 70.66 & 73.08 & 70.67 \\
          & KD+Ours & 76.62 & 74.36 & 74.37 &    76.11   & 71.43 & 74.17 & 71.48 \\
          & $\Delta$ & {\color{red}3.29} & {\color{red}1.38}  & {\color{red}0.83}  & {\color{red}1.19} & {\color{red}0.77}  & {\color{red}1.09}  & {\color{red}0.81}  \\
\cmidrule{2-9}          & KD+CTKD~\cite{li2022curriculum} & 73.39 & 73.52 & 73.93 & 75.45 & 71.19 & 73.52 & 70.99 \\
& KD+CTKD+Ours & 76.67 & 74.47 & 74.58 & 76.08 & 71.34 & 74.01 & 71.39 \\
& $\Delta$ & {\color{red}3.28} & {\color{red}0.95} & {\color{red}0.65} & {\color{red}0.63} & {\color{red}0.15} & {\color{red}0.49} & {\color{red}0.40} \\
\cmidrule{2-9}          & DKD~\cite{zhao2022decoupled} & 76.32   & 74.68 & 74.81 & 76.24 & 71.97 & 74.11 & 71.06 \\
          & DKD+Ours & 77.01 & 74.81 & 74.89   & 76.39  & \underline{72.32} & \underline{74.29} & 71.85 \\
          & $\Delta$  & {\color{red}0.69} & {\color{blue}0.13}  &  {\color{blue}0.08} & {\color{red}0.15}  & {\color{red}0.35}  & {\color{red}0.18}  & {\color{red}0.79} \\
\cmidrule{2-9}     & MLKD~\cite{zhao2022decoupled} & 77.08   & \underline{75.18} &  \underline{75.35} & \underline{76.63}  & 72.19 &  74.11 & \underline{71.89} \\
          & MLKD+Ours & \textbf{78.28} & \textbf{75.22} & \textbf{75.56} &  \textbf{76.95} & \textbf{72.33} & \textbf{74.32} & \textbf{72.27} \\
          & $\Delta$ & {\color{red}1.20} & {\color{blue}0.04} & {\color{red}0.21} & {\color{red}0.32} & {\color{blue}0.14} & {\color{red}0.21} & {\color{red}0.38} \\   
    \bottomrule
    \end{tabular}%
  \label{tab:sameStru}%
  \vspace{-4mm}
\end{table*}%

\begin{table}[tbp]
\footnotesize
\tabcolsep=1mm
\renewcommand\arraystretch{0.9} 
  \centering
  \caption{The top-1 and top-5 accuracy (\%) on the ImageNet validation set~\cite{russakovsky2015imagenet}. 
  The best and second best results are emphasized in \textbf{bold} and \underline{underlined}.
  }
  \vspace{-2mm}
    \begin{tabular}{lllll}
    \toprule
    Teacher/Student & \multicolumn{2}{l}{ResNet34/ResNet18} & \multicolumn{2}{l}{ResNet50/MN-V1} \\
    \midrule
    Accuracy  & top-1 & top-5 & top-1 & top-5 \\
    \midrule
    Teacher & 73.31 & 91.42 & 76.16 & 92.86 \\
    Student & 69.75 & 89.07 & 68.87 & 88.76 \\
    \midrule
    AT~\cite{zagoruyko2017paying}    & 70.69 & 90.01 & 69.56 & 89.33 \\
    OFD~\cite{heo2019comprehensive}   & 70.81 & 89.98 & 71.25 & 90.34 \\
    CRD~\cite{tian2019contrastive}   & 71.17 & 90.13 & 71.37 & 90.41 \\
    ReviewKD~\cite{chen2021distilling} & 71.61 & 90.51 & 72.56 & 91.00 \\
    SimKD~\cite{chen2022knowledge} & 71.59 & 90.48 & 72.25 & 90.86 \\
    CAT-KD~\cite{guo2023class} &  71.26 & 90.45 & 72.24 & 91.13 \\
    \midrule
    KD~\cite{hinton2015distilling}    & 71.03 & 90.05 & 70.50 & 89.80 \\
    KD+Ours & 71.42\(_{\color{red}{\tiny+0.39}}\) & 90.29\(_{\color{red}{\tiny+0.24}}\) & 72.18\(_{\color{red}{\tiny+1.68}}\) & 90.80\(_{\color{red}{\tiny+1.00}}\) \\
    \midrule
    KD+CTKD~\cite{li2022curriculum}  & 71.38 & 90.27 & 71.16 & 90.11 \\
    KD+CTKD+Ours &   71.81\(_{\color{red}{\tiny+0.43}}\)    &  90.46\(_{\color{red}{\tiny+0.19}}\)     & 72.92\(_{\color{red}{\tiny+1.76}}\) & 91.25\(_{\color{red}{\tiny+1.14}}\) \\
    \midrule
    DKD~\cite{zhao2022decoupled}   & 71.70 & 90.41 & 72.05 & 91.05 \\
    DKD+Ours & 71.88\(_{\color{red}{\tiny+0.18}}\) & \underline{90.58}\(_{\color{red}{\tiny+0.17}}\) & 72.85\(_{\color{red}{\tiny+0.80}}\) & 91.23\(_{\color{red}{\tiny+0.18}}\) \\
    \midrule
    MLKD~\cite{jin2023multi}   & \underline{71.90} & 90.55 & \underline{73.01} & \underline{91.42} \\
    MLKD+Ours & \textbf{72.08}\(_{\color{red}{\tiny+0.18}}\) & \textbf{90.74}\(_{\color{red}{\tiny+0.19}}\) & \textbf{73.22}\(_{\color{red}{\tiny+0.21}}\) & \textbf{91.59}\(_{\color{red}{\tiny+0.17}}\) \\
    \bottomrule
    \end{tabular}%
  \label{tab:imagenet}%
\end{table}%


\noindent\textbf{Implementation Details.} We follow the same experimental settings as previous works~\cite{chen2021distilling,zhao2022decoupled,jin2023multi}. 
For the experiments on CIFAR-100, the optimizer is SGD~\cite{sutskever2013importance} and the epoch number is $240$, except for MLKD being 480~\cite{jin2023multi}. 
The learning rate is set initially 0.01 for MobileNets\cite{howard2017mobilenets,sandler2018mobilenetv2} and ShuffleNets~\cite{zhang2018shufflenet} and 0.05 for other architectures consisting of ResNets~\cite{he2016deep}, WRNs~\cite{zagoruyko2016wide} and VGGs~\cite{simonyan2014very}. 
All results are reported by taking average over 4 trials. 
More detailed experimental settings are elaborated in the supplements.

\vspace{-1mm}
\subsection{Main Results}
\vspace{-1mm}

\noindent\textbf{Results on CIFAR-100.} 
We compare the KD results of different methods under various teacher/student settings in Tab.~\ref{tab:diffStru} and \ref{tab:sameStru}. 
Tab.~\ref{tab:diffStru} shows the cases that the teacher and student models have distinct structures, while Tab.~\ref{tab:sameStru} demonstrates the cases that they have the same architecture. 

We evaluate our pre-process on four existing logit-based KD methods. As shown, our {\zscore} standardization can constantly improve their performances. 
After applying our pre-process, the vanilla KD~\cite{hinton2015distilling} achieves comparable performance to the state-of-the-art (SOTA) feature-based methods. 
As SOTA logit-based methods, DKD~\cite{zhao2022decoupled} and MLKD~\cite{jin2023multi} can also be further boosted by our pre-process. 
CTKD~\cite{li2022curriculum} is a KD method that can determine sample-wise temperatures. 
We combine it with ours by leveraging it to predict the base temperature in Algo.~\ref{alg::zscore}. 
As illustrated in tables, student models distilled by CTKD benefits from our pre-process as well. 

\noindent\textbf{Results on ImageNet} of different methods in terms of top-1 and top-5 accuracy are compared in Tab.~\ref{tab:imagenet}. 
Our pre-process can achieve consistent improvement for all the three logit-based methods on the large-scale dataset as well.

\begin{figure}[t]
\begin{center}
   \subfloat{\includegraphics[width=0.333\linewidth]{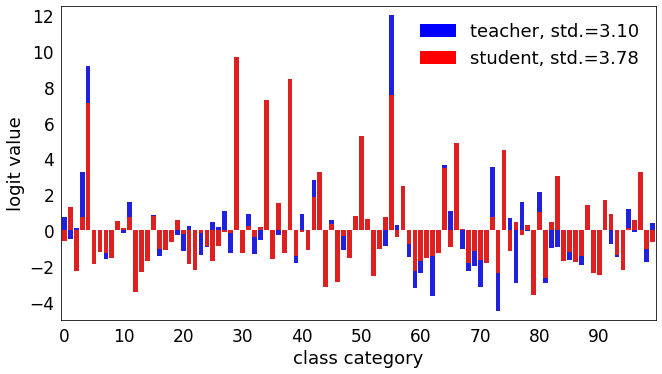}}
   \hfill
   \subfloat{\includegraphics[width=0.333\linewidth]{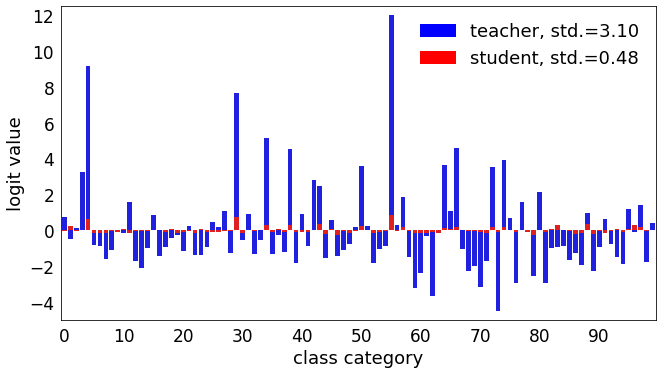}}
   \hfill
   \subfloat{\includegraphics[width=0.333\linewidth]{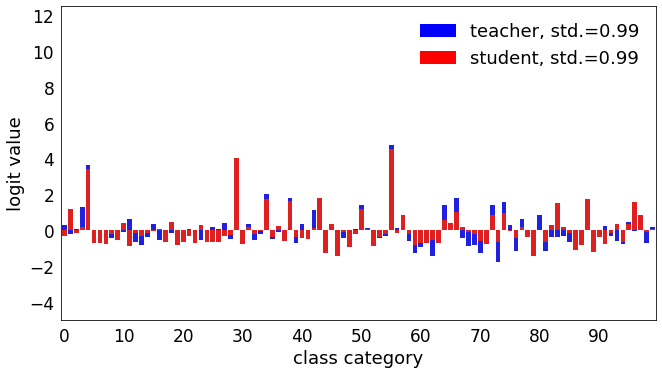}}
   
   \setcounter{subfigure}{0}
   \subfloat[Vanilla KD\\ Mean: 0.27, Max: 3.03.]{\label{fig::diff_kd}\includegraphics[width=0.333\linewidth]{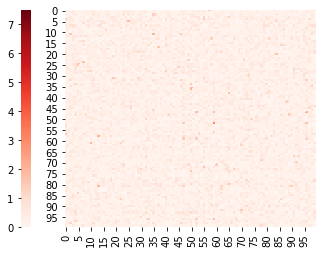}}
   \hfill
   \subfloat[Ours w/o {\zscore}\\ Mean: 0.94, Max: 7.36.]{\label{fig::diff_wo_stand}\includegraphics[width=0.333\linewidth]{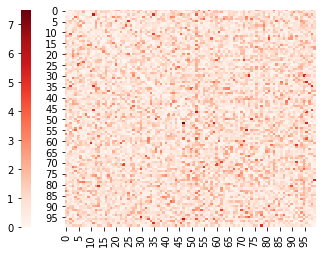}}
   \hfill
   \subfloat[Ours w/ {\zscore}\\ Mean: 0.18, Max:1.18.]{\label{fig::diff_w_stand}\includegraphics[width=0.333\linewidth]{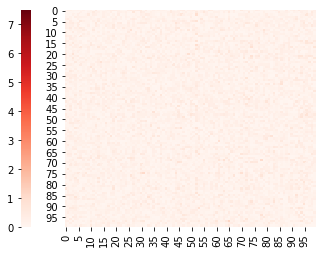}}
\end{center}
\vspace{-6mm}
   \caption{\textbf{1st Row}: An example bar plot of logit output. \textbf{2nd Row}: The heatmap of the average logit difference between the teacher and student. Our pre-process indeed enables the student to generate the logits of divergent range from the teacher as shown in \ref{fig::diff_wo_stand}, while its standardized logits (\ref{fig::diff_w_stand}) are more closer to the teacher's than vanilla KD (\ref{fig::diff_kd}).}
\label{fig:diff}
\end{figure}

\begin{figure*}[t]
\footnotesize
\begin{center}
   \begin{minipage}{0.111\linewidth}
    \centering
    \begin{annotationimage}{width=1\linewidth}{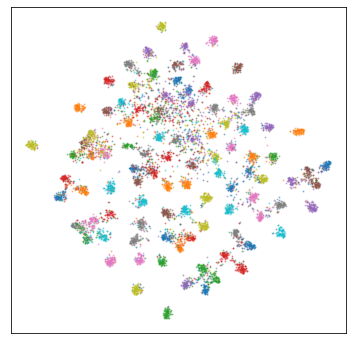}
    \end{annotationimage}
    \subcaption[]{Teacher}
    \end{minipage}
   \hspace{-1.8mm}
  \begin{minipage}{0.222\linewidth}
    \centering
  \begin{subfigure}{0.499\linewidth}
    \begin{annotationimage}{width=1\linewidth}{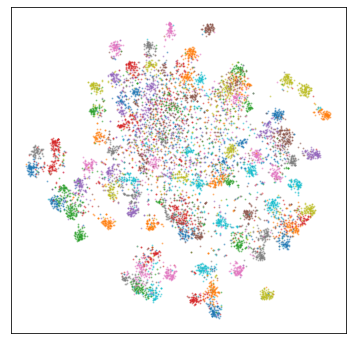}
    \node[fill opacity=1] at (0.31,0.11) {w/o Ours};
    \end{annotationimage}
  \end{subfigure}
    \hspace{-1.6mm}
     \begin{subfigure}{0.499\linewidth}
    \begin{annotationimage}{width=1\linewidth}{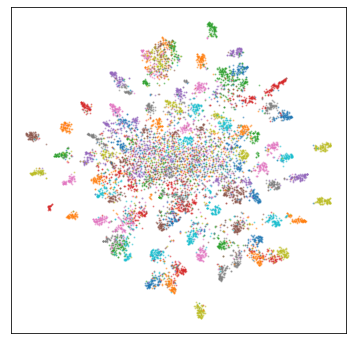}
    \node[fill opacity=1] at (0.28,0.11) {w/ Ours};
    \end{annotationimage}
    \end{subfigure}
    \subcaption[]{KD~\cite{hinton2015distilling}}
    \end{minipage}
   \hspace{-1.8mm}
  \begin{minipage}{0.222\linewidth}
    \centering
  \begin{subfigure}{0.499\linewidth}
    \begin{annotationimage}{width=1\linewidth}{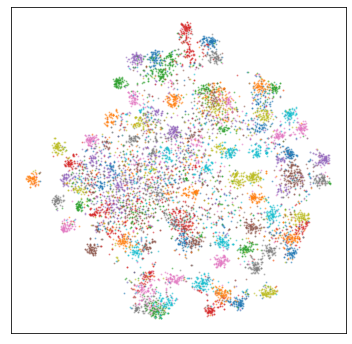}
    \node[fill opacity=1] at (0.31,0.11) {w/o Ours};
    \end{annotationimage}
  \end{subfigure}
    \hspace{-1.6mm}
     \begin{subfigure}{0.499\linewidth}
    \begin{annotationimage}{width=1\linewidth}{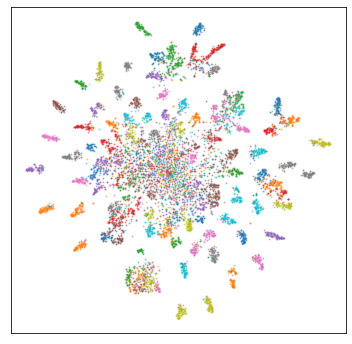}
    \node[fill opacity=1] at (0.28,0.11) {w/ Ours};
    \end{annotationimage}
    \end{subfigure}
    \subcaption[]{CTKD~\cite{li2022curriculum}}
    \end{minipage}
   \hspace{-1.8mm}
  \begin{minipage}{0.222\linewidth}
    \centering
  \begin{subfigure}{0.499\linewidth}
    \begin{annotationimage}{width=1\linewidth}{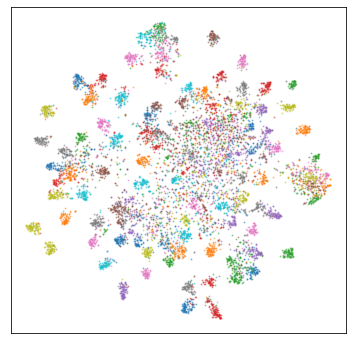}
    \node[fill opacity=1] at (0.31,0.11) {w/o Ours};
    \end{annotationimage}
  \end{subfigure}
    \hspace{-1.6mm}
     \begin{subfigure}{0.499\linewidth}
    \begin{annotationimage}{width=1\linewidth}{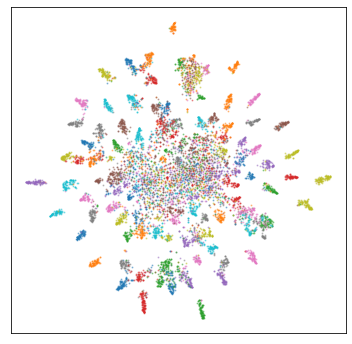}
    \node[fill opacity=1] at (0.28,0.11) {w/ Ours};
    \end{annotationimage}
    \end{subfigure}
    \subcaption[]{DKD~\cite{zhao2022decoupled}}
    \end{minipage}
   \hspace{-1.8mm}
  \begin{minipage}{0.222\linewidth}
    \centering
  \begin{subfigure}{0.499\linewidth}
    \begin{annotationimage}{width=1\linewidth}{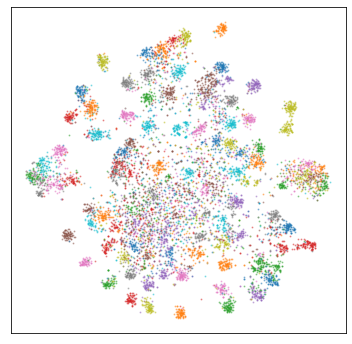}
    \node[fill opacity=1] at (0.31,0.11) {w/o Ours};
    \end{annotationimage}
  \end{subfigure}
    \hspace{-1.6mm}
     \begin{subfigure}{0.499\linewidth}
    \begin{annotationimage}{width=1\linewidth}{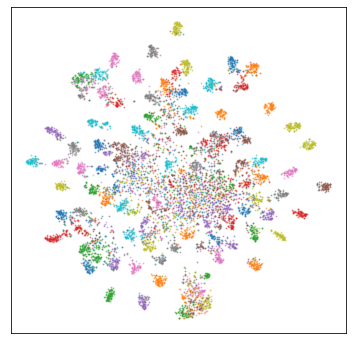}
    \node[fill opacity=1] at (0.28,0.11) {w/ Ours};
    \end{annotationimage}
    \end{subfigure}
    \subcaption[]{MLKD~\cite{jin2023multi}}
    \end{minipage}
\end{center}
\vspace{-7mm}
   \caption{The t-SNE~\cite{van2008visualizing} visualization of features. The teacher and student are ResNet32$\times$4 and ResNet8$\times$4.}
\label{fig:tsne}
\vspace{-4mm}
\end{figure*}


\noindent\textbf{Ablation Studies}
We conduct extensive ablation studies in terms of different configurations of base temperature and the weight of KD loss $\lambda_{\rm KD}$. 
The part of results when base temperature is 2 are shown in Tab.~\ref{tab:ablation-1}. We can see as the weight of KD loss increases, the vanilla KD where \softmax takes the original logit vectors as input cannot have a considerable performance gain. 
In contrast, our {\zscore} pre-process achieves a significant enhancement. 
More ablation studies of other settings are in the supplements. 


The weight of KD loss is set relatively larger than that of CE loss because our pre-process enables student to focus more on the dark knowledge from teacher instead of hard labels. 
Another reason of the large KD weight is to compensate the gradient involving {\zscore}. 


\subsection{Extensions}
\vspace{-2mm}

\noindent\textbf{Logit range.} 
We plot a bar plot in the first row of Fig.~\ref{fig:diff} to show an example of logit range. 
The second row of Fig.~\ref{fig:diff} illustrates the extent of average logit difference between teacher and student. 
Without applying our pre-process, the student fails to produce as large logit as the teacher at the target label index (7.5 v.s. 12). 
The mean average distance between the teacher and student logits also reaches 0.27. 
The restriction of logit range impedes the student to predict correctly. 
In contrast, our pre-process breaks the restriction and enables student to generate the logits of appropriate range for itself. 
Its effective logit output after standardization on the contrary matches the teacher's nicely. 
The mean distance of standardized logits also shrinks to 0.18, implying its better mimicking the teacher.

\noindent\textbf{Logit variance.} 
As illustrated in the first row of Fig.~\ref{fig:diff}, the vanilla KD forces the variance of the student logits approaches the teacher's (3.78 v.s. 3.10). 
However, our pre-process breaks the shackle and the student logit can have flexible logit variance (0.48 v.s. 3.10), while its standardized logits have the same variance as the teacher (both 0.99).

\noindent\textbf{Feature visualizations} of deep representations are shown in Fig.~\ref{fig:tsne}. 
As implied, our pre-process improves the feature separability and discriminability of all the methods including KD~\cite{hinton2015distilling}, CTKD~\cite{li2022curriculum}, DKD~\cite{zhao2022decoupled} and MLKD~\cite{jin2023multi}.

\begin{figure}[t]
\begin{center}
   \subfloat[Teacher models of different sizes]{\label{fig:b_hist-a}\includegraphics[width=1.\linewidth]{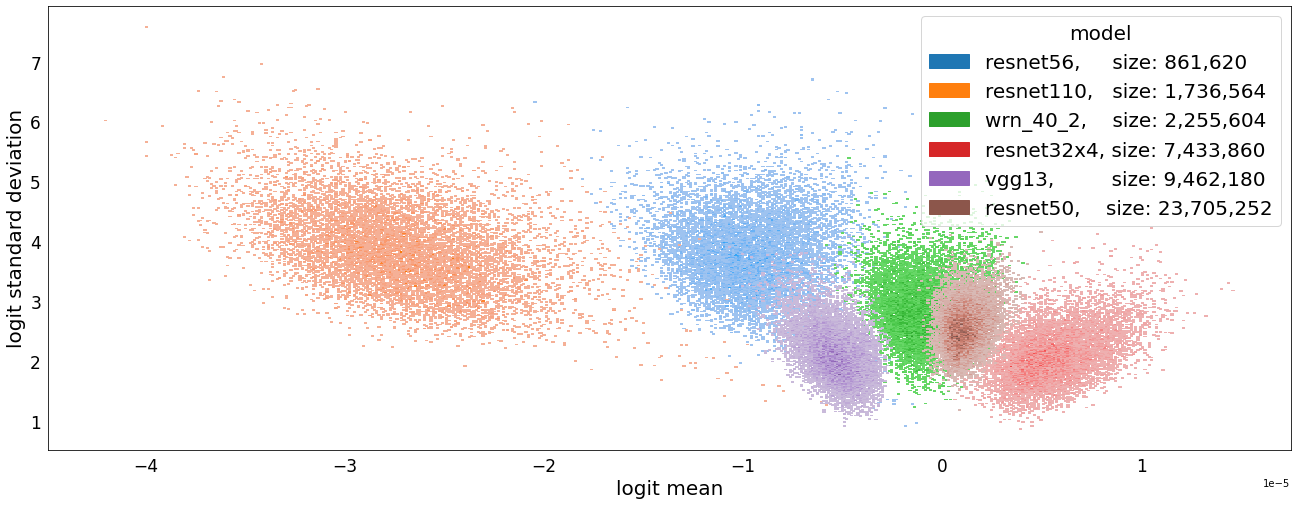}}


   \subfloat[Teacher and student distilled by KD only and KD with our pre-process]{\label{fig:b_hist-b}\includegraphics[width=1.\linewidth]{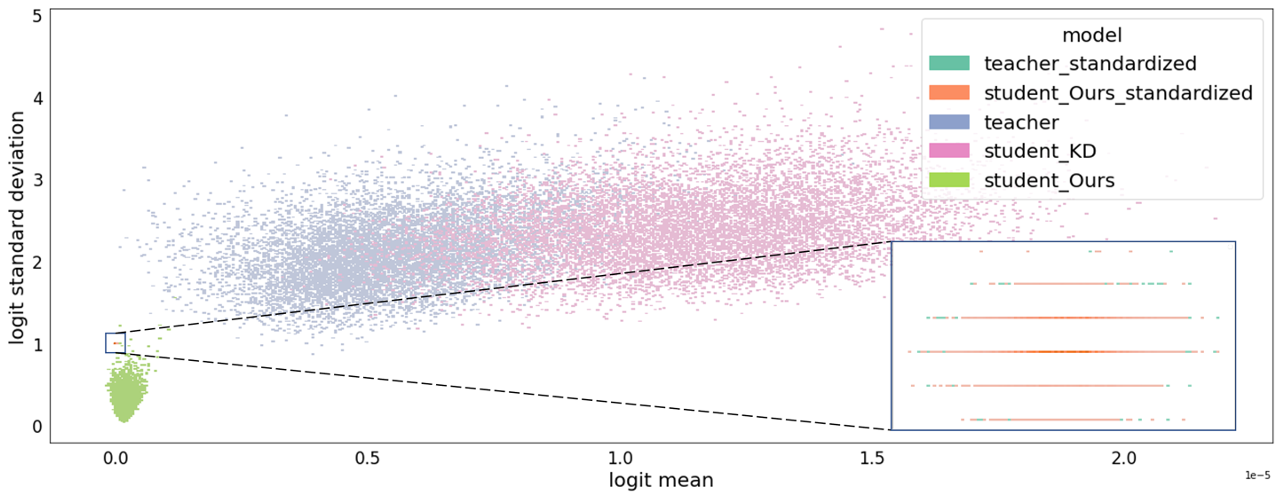}}
\end{center}
\vspace{-6mm}
   \caption{The bivariate histogram of logit mean and logit standard deviation for multiple models on CIFAR-100. }
\label{fig:b_hist}
\end{figure}

\noindent\textbf{Improving distillation for large teacher.} 
Existing works~\cite{cho2019efficacy,wang2021knowledge,zhao2022decoupled} observe that a larger teacher is unnecessarily a better one. 
The phenomenon implies that the transfer of knowledge from a large teacher is not always as smooth as a small one. 
They explain the observation with the existence of a capacity gap between cumbersome teachers and lightweight students. 
We interpret the issue as the difficulty of students in mimicking logits of comparable range and variance as teachers. 
As shown in the bivariate histogram of Fig.~\ref{fig:b_hist-a}, the bigger teachers like ResNet50 and VGG13 generates more condensed logits of mean closer to zero and smaller standard deviation. 
The tendency shows that smaller models intrinsically predicts the outputs of larger bias from zero mean and larger variance. 
Therefore, when students are small, the same tendency remains and they are difficult to produce as compact logits as large teachers.

We alleviate the problem by our pre-process. 
As shown in Tab.~\ref{tab:various_tea}, our pre-process consistently improves the distillation performance for various teachers of different sizes and capacities. 
We also show the mimicking goodness of students by a bivariate histogram plot in Fig.~\ref{fig:b_hist-b}. 
The student distilled by vanilla KD predicts logits apparently deviated from teachers' in terms of logit mean and standard deviation. 
In contrast, our pre-process enables the student to perfectly matching the teacher in terms of standardized logit mean and standard deviation (see the zoomed-in region at bottom right corner).

\noindent\textbf{More experiments} of distilling vision Transformers~\cite{Touvron2021Training,wu2022tiny_vit,Chen2022DearKD,li2022Locality,li2023Automated} 
are attached in the supplements.




\begin{table}[tbp]
\footnotesize
\tabcolsep=0.0mm
\renewcommand\arraystretch{0.8} 
  \centering
  \caption{The ablation studies under different settings in our {\zscore}. The base temperature $\tau$ is set to be 2. By default $\lambda_{\rm CE}=0.1$. The logit vector of teacher $\vv_n$ and student $\zv_n$ are abbreviated as $\zv$ for succinctness. The teacher and student are ResNet32$\times$4 and ResNet8$\times$4.}
  \vspace{-3mm}
    \begin{tabular}{c|cccc}
    \toprule
    \makebox[0.07\textwidth][c]{$\lambda_{\rm KD}$} & \makebox[0.09\textwidth][c]{$\zv$ (KD)}     & \makebox[0.09\textwidth][c]{$\zv-\overline \zv$} & \makebox[0.09\textwidth][c]{$\frac{\zv}{\sigma(\zv)}$} & \makebox[0.09\textwidth][c]{$\frac{(\zv-\overline \zv)}{\sigma(\zv)}$ (Ours)} \\
    \midrule
    0.9     & 73.60 & 73.37 & 73.79 & 74.14 \\
    3.0     & 74.38 & 74.33 & 75.86 & 76.11 \\
    6.0     & 74.45 & 74.82 & 76.44 & 76.56 \\
    9.0     & 73.33 & 73.94 & 76.30 & 76.62 \\
    12.0    & 68.29 & 71.56 & 76.49 & 76.56 \\
    15.0    & 65.34 & 62.01 & 76.42 & 76.61 \\
    18.0    & 63.45 & 61.31 & 76.18 & 76.33 \\
    \bottomrule
    \end{tabular}%
  \label{tab:ablation-1}%
  \vspace{2mm}
\end{table}%

\begin{table}[tbp]
\footnotesize
\tabcolsep=0.2mm
\renewcommand\arraystretch{0.9} 
  \centering
  \caption{The results of distillation of various teacher models on CIFAR-100. The student model is WRN-16-2. 
  }
  \vspace{-3mm}
    \begin{tabular}{lcccccc}
    \toprule
    \multirow{2}{*}{Teacher} & \makebox[0.065\textwidth][c]{VGG13} & \makebox[0.065\textwidth][c]{W-28-2} & \makebox[0.065\textwidth][c]{W-40-2} & \makebox[0.065\textwidth][c]{W-16-4} & \makebox[0.065\textwidth][c]{W-28-4} & \makebox[0.065\textwidth][c]{ResNet50} \\
          & 74.64 & 75.45 & 75.61 & 77.51 & 78.60 & 79.34 \\
    \midrule
    KD~\cite{hinton2015distilling}    & 74.93 & 75.37 & 74.92 & 75.79 & 75.04 & 75.36 \\
    KD+Ours & 75.03 & 76.32 & 76.11 & 76.72 & 75.77 & 76.24 \\
    \midrule
    DKD~\cite{zhao2022decoupled}   & 75.45 & 75.92 & 76.24 & 76.00 & 76.45 & 76.60 \\
    DKD+Ours & 75.56 & 76.39 & 76.39 & 76.68 & 76.67 & 76.82 \\
    \bottomrule
    \end{tabular}%
  \label{tab:various_tea}%
\end{table}%

\vspace{-1.5mm}
\section{Conclusion}
\vspace{-1.5mm}
In this work, we identify a lack of theoretical support for the global and shared temperature in conventional KD pipelines. 
Our analysis based on the principle of entropy maximization leads to the conclusion that the temperature is derived from a flexible Lagrangian multiplier, allowing for a flexible value assignment. 
We then highlight several drawbacks associated with the conventional practice of sharing temperatures between teachers and students. Additionally, 
we presented a toy case where the KD pipeline with shared temperature led to an inauthentic evaluation of student performance.
To mitigate the concerns, we propose a logit {\zscore} standardization as a pre-process to enable student to focus on the innate relations of teacher logits rather than logit magnitude. 
The extensive experiments demonstrate the effectiveness of our pre-process in enhancing the existing logit-based KD methods.


\noindent\textbf{Acknowledgements}. This work has been supported in part by National Natural Science Foundation of China (No. 62322216, 62025604, 62306308), in part by Shenzhen Science and Technology Program (Grant No. JCYJ20220818102012025, KQTD20221101093559018).

{
    \small
    \bibliographystyle{ieeenat_fullname}
    \bibliography{main}
}


\end{document}